\documentclass[10pt,twocolumn,letterpaper]{article}

\usepackage[pagenumbers]{cvpr} 

\usepackage{graphicx}
\usepackage{amsmath}
\usepackage{amssymb}
\usepackage{booktabs}

\usepackage{blindtext}
\usepackage{multirow}
\usepackage{bm}
\usepackage{bbm} 
\usepackage{comment}
\usepackage{algorithm}
\usepackage{algorithmic}
\usepackage{array} 
\usepackage[dvipsnames]{xcolor}
\usepackage[safe]{tipa} 
\usepackage{color, colortbl}
\usepackage[misc,geometry]{ifsym} 

\makeatletter
\renewcommand\paragraph{
  \@startsection{paragraph} 
  {4} 
  {\z@} 
  {.5em \@plus1ex \@minus.2ex} 
  {-1.5em} 
  {\normalfont\normalsize\bfseries} 
}
\makeatother

\makeatletter
\def\@fnsymbol#1{\ensuremath{\ifcase#1\or \textsuperscript{~\Letter}\or \ddagger\or
   \mathsection\or \mathparagraph\or \|\or **\or \dagger\dagger
   \or \ddagger\ddagger \else\@ctrerr\fi}}
\makeatother

\newcommand{\tableCellHeight}{1}
\newcommand{\tabstyle}[1]{
  \setlength{\tabcolsep}{#1}
  \renewcommand{\arraystretch}{\tableCellHeight}
  \centering
  \small
}

\newcommand{\romannum}[1]{\romannumeral #1} 
\newcommand{\rotbox}[1]{\rotatebox{90}{#1}}

\newcommand{\hgreen}[1]{\textcolor{ForestGreen}{\textbf{#1}}} 
\newcommand{\hblue}[1]{\textcolor{NavyBlue}{\textbf{#1}}} 
\definecolor{tabhighlight}{HTML}{e5e5e5}
\definecolor{citecolor}{HTML}{0071bc}

\usepackage[pagebackref,breaklinks,colorlinks,citecolor=citecolor]{hyperref}

\usepackage[capitalize]{cleveref}
\crefname{section}{Sec.}{Secs.}
\Crefname{section}{Section}{Sections}
\Crefname{table}{Table}{Tables}
\crefname{table}{Tab.}{Tabs.}

\def\cvprPaperID{2629} 
\def\confName{CVPR}
\def\confYear{2022}

\begin{document}

\title{Conditional Prompt Learning for Vision-Language Models}

\author{
Kaiyang Zhou \and
Jingkang Yang \and
Chen Change Loy \and
Ziwei Liu\thanks{Corresponding author}
\and
S-Lab, Nanyang Technological University, Singapore \\
{\tt\small \{kaiyang.zhou, jingkang001, ccloy, ziwei.liu\}@ntu.edu.sg}
}
\maketitle

\begin{abstract}
With the rise of powerful pre-trained vision-language models like CLIP, it becomes essential to investigate ways to adapt these models to downstream datasets. A recently proposed method named Context Optimization (CoOp) introduces the concept of prompt learning---a recent trend in NLP---to the vision domain for adapting pre-trained vision-language models. Specifically, CoOp turns context words in a prompt into a set of learnable vectors and, with only a few labeled images for learning, can achieve huge improvements over intensively-tuned manual prompts. In our study we identify a critical problem of CoOp: the learned context is not generalizable to wider unseen classes within the same dataset, suggesting that CoOp overfits base classes observed during training. To address the problem, we propose Conditional Context Optimization (CoCoOp), which extends CoOp by further learning a lightweight neural network to generate for each image an input-conditional token (vector). Compared to CoOp's static prompts, our dynamic prompts adapt to each instance and are thus less sensitive to class shift. Extensive experiments show that CoCoOp generalizes much better than CoOp to unseen classes, even showing promising transferability beyond a single dataset; and yields stronger domain generalization performance as well. Code is available at \url{https://github.com/KaiyangZhou/CoOp}.
\end{abstract}


\section{Introduction}
\label{sec:intro}

\begin{figure*}[t]
    \centering
    \begin{subfigure}[b]{\textwidth}
    \centering
    \includegraphics[width=\textwidth]{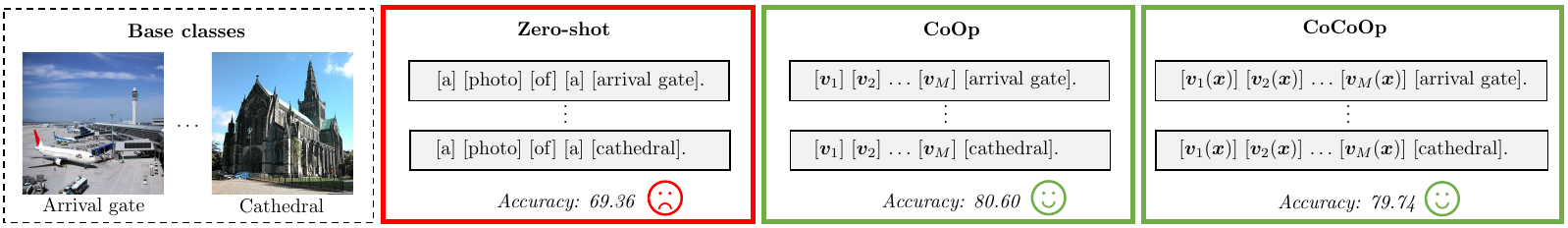}
    \caption{Both CoOp and CoCoOp work well on the base classes observed during training and beat manual prompts by a significant margin.}
    \label{fig:motivation01}
    \end{subfigure}
    ~
    \begin{subfigure}[b]{\textwidth}
    \centering
    \includegraphics[width=\textwidth]{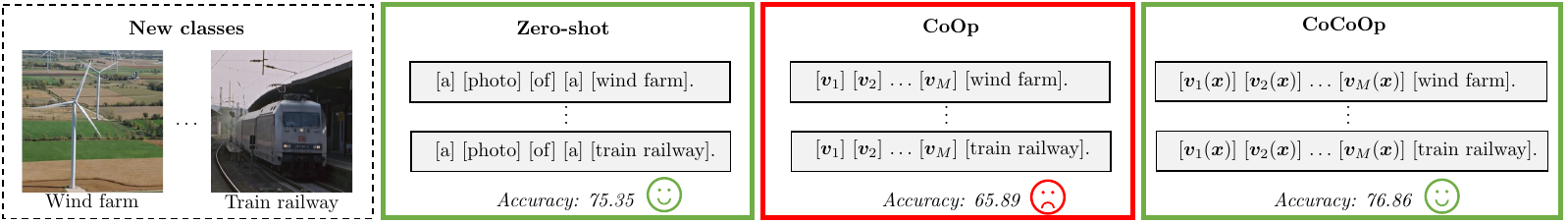}
    \caption{The instance-conditional prompts learned by CoCoOp are much more generalizable than CoOp to the unseen classes.}
    \label{fig:motivation02}
    \end{subfigure}
    \caption{\textbf{Motivation of our research: to learn generalizable prompts}. The images are randomly selected from SUN397~\cite{xiao2010sun}, which is a widely-used scene recognition dataset.
    }
    \label{fig:motivation}
\end{figure*}

Recent research in large-scale vision-language pre-training has achieved striking performance in zero-shot image recognition~\cite{radford2021learning,jia2021scaling,furst2021cloob,li2021supervision}, demonstrating a potential in learning open-world visual concepts for such a paradigm. The key design lies in how visual concepts are modeled. In traditional supervised learning where labels are discretized, each category is associated with a randomly initialized weight vector that is learned to minimize the distance with images containing the same category. Such a learning method focuses on closed-set visual concepts, limiting the model to a pre-defined list of categories, and is unscalable when it comes to new categories unseen during training.

In contrast, for vision-language models\footnote{We follow existing studies~\cite{radford2021learning,jia2021scaling,furst2021cloob,li2021supervision} to refer to CLIP-like models as \emph{vision-language models}.} like CLIP~\cite{radford2021learning} and ALIGN~\cite{jia2021scaling}, the classification weights are diametrically generated by a parameterized text encoder (e.g., a Transformer~\cite{vaswani2017attention}) through prompting~\cite{liu2021pre}. For instance, to differentiate pet images containing different breeds of dogs and cats, one can adopt a prompt template like ``a photo of a \{class\}, a type of pet'' as input to the text encoder, and as a result, class-specific weights for classification can be synthesized by filling in the ``\{class\}'' token with real class names. Compared to discrete labels, vision-language models' source of supervision comes from natural language, which allows open-set visual concepts to be broadly explored and has been proven effective in learning transferable representations~\cite{radford2021learning,jia2021scaling}.

With the rise of such powerful vision-language models, the community has recently started to investigate potential solutions to efficiently adapt these models to downstream datasets~\cite{zhou2021coop,yao2021cpt,gao2021clip,wortsman2021robust}. To fit web-scale data, such as the 400 million pairs of images and texts used by CLIP, vision-language models are purposefully designed to have high capacity, entailing that the model size would be enormous, typically with hundreds of millions of parameters or even billions. Therefore, fine-tuning the entire model, as often adopted in deep learning research~\cite{he2016deep}, is impractical and might even damage the well-learned representation space.

A safer approach is to tune a prompt by adding some context that is meaningful to a task, like ``a type of pet'' for the pet dataset mentioned above, which has been found effective in improving performance~\cite{radford2021learning}. However, prompt engineering is extremely time-consuming and inefficient as it has to be based on trial and error, and does not guarantee an optimal prompt either. To automate prompt engineering, Zhou et al.~\cite{zhou2021coop} have recently explored the concept of prompt learning---a recent trend in NLP~\cite{shin2020autoprompt,jiang2020can,li2021prefix,zhong2021factual,lester2021power,gao2020making}---for adapting pre-trained vision-language models. Their approach, Context Optimization (CoOp), turns context words in a prompt into a set of learnable vectors, taking advantage of the differentiable nature of neural networks. With only a few labeled images for learning, CoOp achieves huge improvements over intensively-tuned manual prompts across a wide range of image recognition datasets.

In our study, we identify a critical problem of CoOp: the learned context is not generalizable to wider unseen classes within the same task. Figure~\ref{fig:motivation} illustrates the problem: the context learned by CoOp works well in distinguishing the base classes like ``arrival gate'' and ``cathedral'' but suffers a significant drop in accuracy when it is transferred to the new (unseen) classes, such as ``wind farm'' and ``train railway''---even though the task's nature remains the same, i.e., recognizing scenes. The results suggest that the learned context overfits the base classes, thus failing to capture more generalizable elements that are vital for broader scene recognition. We argue that such a problem is caused by CoOp's static design: the context, which is fixed once learned, is optimized only for a specific set of (training) classes. On the contrary, the manually-designed prompts adopted by the zero-shot method are relatively generalizable.

\begin{figure}[t]
    \centering
    \includegraphics[width=\columnwidth]{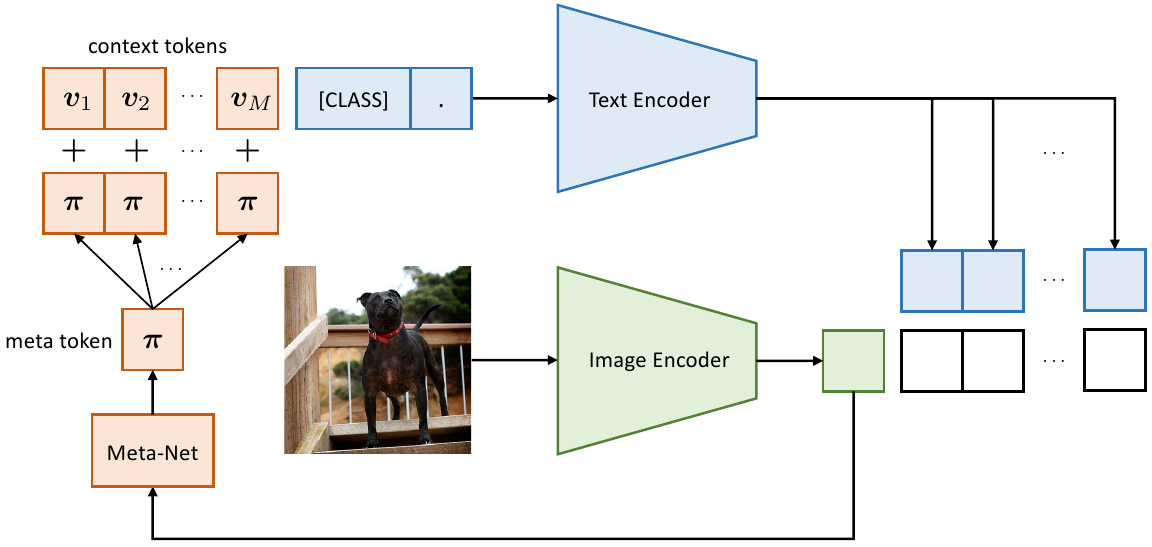}
    \caption{Our approach, Conditional Context Optimization (CoCoOp), consists of two learnable components: a set of context vectors and a lightweight neural network (Meta-Net) that generates for each image an input-conditional token.}
    \label{fig:method}
\end{figure}

To address the weak generalizability problem, we introduce a novel concept: \emph{conditional prompt learning}. The key idea is to make a prompt conditioned on each input instance (image) rather than fixed once learned. To make the model parameter-efficient, we introduce a simple yet effective implementation of conditional prompt learning. Specifically, we extend CoOp by further learning a lightweight neural network to generate for each image an input-conditional token (vector), which is combined with the learnable context vectors. We call our approach Conditional Context Optimization (CoCoOp).\footnote{Pronounced as /k\textschwa\textupsilon\textsecstress ku:p/.} An overview is shown in Figure~\ref{fig:method}. Interestingly, the paradigm of CoCoOp is analogous to image captioning~\cite{vinyals2015show}, which explains why instance-conditional prompts are more generalizable: \emph{they are optimized to characterize each instance (more robust to class shift) rather than to serve only for some specific classes}.

We present comprehensive experiments on 11 datasets covering a diverse set of visual recognition tasks. Specifically, we design a base-to-new generalization setting where a model is first learned using base classes and then tested on completely new classes. Compared with the zero-shot method~\cite{radford2021learning} and CoOp~\cite{zhou2021coop}, our approach achieves the best overall performance (Table~\ref{tab:results_generalization}). Importantly, CoCoOp gains significant improvements over CoOp in unseen classes (Figure~\ref{fig:coop_vs_ours_base2new}(a)), allowing the gap between manual and learning-based prompts to be substantially reduced.

In a more challenging scenario where the context learned for one task is directly transferred to other tasks with drastically different classes, CoCoOp still beats CoOp with a clear margin (Table~\ref{tab:xd}), suggesting that instance-conditional prompts are more transferable and have the potential to succeed at larger scale. CoCoOp also obtains stronger domain generalization performance than CoOp (Table~\ref{tab:dg}), further justifying the strengths of dynamic prompts.

In summary, our research provides timely insights into the generalizability problem in prompt learning, and crucially, demonstrates the effectiveness of a simple idea in various problem scenarios. We hope our approach and the findings presented in this work can pave the way for future research in generalizable---and transferable---prompt learning.

\section{Related Work}
\label{sec:related_work}

\paragraph{Vision-Language Models}
We mainly review studies focused on aligning images and texts to learn a joint embedding space~\cite{radford2021learning,jia2021scaling,zhang2020contrastive}. The idea of cross-modality alignment is certainly not new and has been investigated since nearly a decade ago---though with dramatically different technologies than today.

A typical vision-language model consists of three key elements: two for image and text encoding while the third is related to the design of loss functions. In early days, models for processing images and texts are often designed and also learned independently, with their outputs connected by extra modules (losses) for alignment. Images are often encoded using hand-crafted descriptors~\cite{socher2013zero,elhoseiny2013write} or neural networks~\cite{frome2013devise,lei2015predicting}, while texts are encoded using, for instance, pre-trained word vectors~\cite{socher2013zero,frome2013devise} or the frequency-based TF-IDF features~\cite{elhoseiny2013write,lei2015predicting}. In terms of cross-modality alignment, common approaches include metric learning~\cite{frome2013devise}, multi-label classification~\cite{joulin2016learning,gomez2017self}, and n-gram language learning~\cite{li2017learning}. Recently, a study suggests that training the vision part with an image captioning loss can make the visual representation more transferable~\cite{desai2021virtex}.

Recent vision-language models~\cite{radford2021learning,jia2021scaling,furst2021cloob,li2021supervision} bridge the two modalities by learning two encoders jointly. Also, the models are now built with much larger neural networks. As discussed in Zhou et al.~\cite{zhou2021coop}, recent successes in vision-language models are mainly attributed to the developments in \romannum{1}) Transformers~\cite{vaswani2017attention}, \romannum{2}) contrastive representation learning~\cite{chen2020simple,he2020momentum,henaff2020data}, and \romannum{3}) web-scale training datasets~\cite{radford2021learning,jia2021scaling}. A representative approach is CLIP~\cite{radford2021learning}, which trains two neural network-based encoders using a contrastive loss to match pairs of images and texts. After consuming 400 million data pairs, the CLIP model demonstrates a remarkable zero-shot image recognition capability. Similar to CoOp~\cite{zhou2021coop}, our approach is orthogonal to the research of CLIP-like models~\cite{radford2021learning,jia2021scaling,furst2021cloob,li2021supervision}, aiming to offer an efficient solution for adapting pre-trained vision-language models to downstream applications.

\paragraph{Prompt Learning}
This topic originates from the NLP domain. The motivation was to view pre-trained language models, such as BERT~\cite{devlin2019bert} or GPT~\cite{radford2019language}, as knowledge bases from which information useful to downstream tasks is elicited~\cite{petroni2019language}. Concretely, given a pre-trained language model, the task is often formulated as a ``fill-in-the-blank'' cloze test, such as asking the model to predict the masked token in ``No reason to watch. \underline{It was} [MASK]'' as either ``positive'' or ``negative'' for sentiment classification. The key lies in how to design the underlined part, known as prompt (template), in such a format familiar to the model.

Instead of manually designing a prompt, research in prompt learning aims to automate the process with the help of affordable-sized labeled data. Jiang et al.~\cite{jiang2020can} use text mining and paraphrasing to generate a group of candidate prompts, within which the optimal ones are chosen to have the highest training accuracy. Shin et al.~\cite{shin2020autoprompt} propose AutoPrompt, a gradient-based approach that selects from a vocabulary the best tokens that cause the greatest changes in gradients based on the label likelihood. Our research is most related to continuous prompt learning methods~\cite{zhong2021factual,li2021prefix,lester2021power}, where the main idea is to turn a prompt into a set of continuous vectors that can be end-to-end optimized with respect to an objective function. See Liu et al.~\cite{liu2021pre} for a more comprehensive survey.

In computer vision, prompt learning is a nascent research direction that has only been explored very recently~\cite{zhou2021coop,yao2021cpt,rao2022denseclip,ju2021prompting,zhang2021pointclip}. Our research is built on top of CoOp~\cite{zhou2021coop}, which is the earliest work to bring continuous prompt learning to the vision domain for adaptation of pre-trained vision-language models. Crucially, our approach solves the weak generalizability problem of CoOp~\cite{zhou2021coop}, based on a simple idea of conditional prompt learning---\emph{which to our knowledge is also novel in the context of NLP and thus could be of interest to the NLP community as well}.

\paragraph{Zero-Shot Learning (ZSL)}
is another relevant research area where the goal is similar to ours, i.e., to recognize novel classes by training only on base classes~\cite{wang2019survey,xian2017zero,chao2016empirical,yi2022exploring}. Moreover, the generalization problem where a model trained on base classes often fails on novel classes is also linked to the ``seen-class bias'' issue raised in the ZSL literature~\cite{xian2017zero}. The most common approach to ZSL is to learn a semantic space based on auxiliary information such as attributes~\cite{huynh2020fine} or word embeddings~\cite{frome2013devise,wang2018zero}. Different from existing ZSL methods, our work addresses the emerging problem of adapting large vision-language models and uses drastically different techniques based on prompting.

\section{Methodology}
\label{sec:methodology}

An overview of our approach is shown in Figure~\ref{fig:method}. Below we first provide brief reviews on CLIP~\cite{radford2021learning}, which is the base model used in this paper, and CoOp~\cite{zhou2021coop}. Then, we present the technical details of our approach as well as the rationale behind the design. Same as CoOp, our approach is applicable to broader CLIP-like vision-language models.

\subsection{Reviews of CLIP and CoOp}
\paragraph{Contrastive Language-Image Pre-training}
known as CLIP~\cite{radford2019language}, has well demonstrated the potential of learning open-set visual concepts. CLIP is built using two encoders, one for image and the other for text, as shown in Figure~\ref{fig:method}. The image encoder can be either a ResNet~\cite{he2016deep} or a ViT~\cite{dosovitskiy2021image}, which is used to transform an image into a feature vector. The text encoder is a Transformer~\cite{vaswani2017attention}, which takes as input a sequence of word tokens and again produces a vectorized representation.

During training, CLIP adopts a contrastive loss to learn a joint embedding space for the two modalities. Specifically, for a mini-batch of image-text pairs, CLIP maximizes for each image the cosine similarity with the matched text while minimizes the cosine similarities with all other unmatched texts, and the loss is computed in a similar fashion for each text too. After training, CLIP can be used for zero-shot image recognition. Let $\bm{x}$ be image features generated by the image encoder and $\{\bm{w}_i\}_{i=1}^K$ a set of weight vectors produced by the text encoder, each representing a category (suppose there are $K$ categories in total). In particular, each $\bm{w}_i$ is derived from a prompt, such as ``a photo of a \{class\}'' where the ``\{class\}'' token is filled with the $i$-th class name. The prediction probability is then
\begin{equation} \label{eq:pred_zs}
p(y | \bm{x}) = \frac{\exp (\operatorname{sim} (\bm{x}, \bm{w}_y) / \tau)}{\sum_{i=1}^K \exp (\operatorname{sim} (\bm{x}, \bm{w}_i) / \tau)},
\end{equation}
where $\operatorname{sim}(\cdot, \cdot)$ denotes cosine similarity and $\tau$ is a learned temperature parameter.

\paragraph{Context Optimization (CoOp)}
aims to overcome the inefficiency problem in prompt engineering for better adapting pre-trained vision-language models to downstream applications~\cite{zhou2021coop}. The key idea in CoOp is to model each context token using a continuous vector that can be end-to-end learned from data. Concretely, instead of using ``a photo of a'' as the context, CoOp introduces $M$ learnable context vectors, $\{\bm{v}_1, \bm{v}_2, \hdots, \bm{v}_M\}$, each having the same dimension with the word embeddings. The prompt for the $i$-th class, denoted by $\bm{t}_i$, now becomes $\bm{t}_i =\{\bm{v}_1, \bm{v}_2, \hdots, \bm{v}_M, \bm{c}_i\}$ where $\bm{c}_i$ is the word embedding(s) for the class name. The context vectors are shared among all classes.\footnote{CoOp has an alternative version that learns class-specific context, which is not considered here because it is not straightforward to transfer class-specific context to unseen classes.} Let $g(\cdot)$ denote the text encoder, the prediction probability is then
\begin{equation} \label{eq:pred_coop}
p(y | \bm{x}) = \frac{\exp (\operatorname{sim} (\bm{x}, g(\bm{t}_y)) / \tau)}{\sum_{i=1}^K \exp (\operatorname{sim} (\bm{x}, g(\bm{t}_i) / \tau)}.
\end{equation}

To adapt CLIP to a downstream image recognition dataset, a cross-entropy loss can be used as the learning objective. Since the text encoder $g(\cdot)$ is differentiable, gradients can be propagated all the way back to update the context vectors. Note that the base model of CLIP is frozen in the entire training process (ours too).

\subsection{CoCoOp: Conditional Context Optimization}
CoOp is a data-efficient approach allowing the context vectors to be trained with only a few labeled images in a downstream dataset. However, as discussed CoOp is not generalizable to wider unseen classes within the same task. We argue that instance-conditional context can generalize better because it shifts the focus away from a specific set of classes---for reducing overfitting---to each input instance, and hence to the entire task.

A straightforward way to implement CoCoOp is to build $M$ neural networks to get $M$ context tokens. However, such a design would require $M \times$ the size of a neural network, which is much larger than having $M$ context vectors as in CoOp. Here we propose a parameter-efficient design that works very well in practice. Specifically, on top of the $M$ context vectors, we further learn a lightweight neural network, called Meta-Net, to generate for each input a conditional token (vector), which is then combined with the context vectors. See Figure~\ref{fig:method} for a sketch of the architecture.

Let $h_{\bm{\theta}}(\cdot)$ denote the Meta-Net parameterized by $\bm{\theta}$, each context token is now obtained by $\bm{v}_m (\bm{x}) = \bm{v}_m + \bm{\pi}$ where $\bm{\pi} = h_{\bm{\theta}} (\bm{x})$ and $m \in \{1, 2, ..., M\}$. The prompt for the $i$-th class is thus conditioned on the input, i.e., $\bm{t}_i (\bm{x}) =\{\bm{v}_1 (\bm{x}), \bm{v}_2 (\bm{x}), \hdots, \bm{v}_M (\bm{x}), \bm{c}_i\}$. The prediction probability is computed as
\begin{equation} \label{eq:pred_cocoop}
p(y | \bm{x}) = \frac{\exp (\operatorname{sim} (\bm{x}, g(\bm{t}_y (\bm{x}))) / \tau)}{\sum_{i=1}^K \exp (\operatorname{sim} (\bm{x}, g(\bm{t}_i (\bm{x})) / \tau)}.
\end{equation}

During training, we update the context vectors $\{\bm{v}_m\}_{m=1}^M$ together with the Meta-Net's parameters $\bm{\theta}$. In this work, the Meta-Net is built with a two-layer bottleneck structure (Linear-ReLU-Linear), with the hidden layer reducing the input dimension by $16 \times$. The input to the Meta-Net is simply the output features produced by the image encoder. We leave exploration of more advanced designs for future work.

\section{Experiments}
\label{sec:experiments}


\begin{table*}[t]
    \tabstyle{6pt}
    \caption{\textbf{Comparison of CLIP, CoOp and CoCoOp in the base-to-new generalization setting}. For learning-based methods (CoOp and CoCoOp), their prompts are learned from the base classes (16 shots). The results strongly justify the strong generalizability of conditional prompt learning. H: Harmonic mean (to highlight the generalization trade-off~\cite{xian2017zero}).}
    \label{tab:results_generalization}
    \begin{subtable}[t]{.3\textwidth}
    \centering
    \caption{\textbf{Average over 11 datasets}.}
    \begin{tabular}{l cc|c}
    \toprule
    & Base & New & H \\
    \midrule
    CLIP & 69.34 & \textbf{74.22} & 71.70 \\
    CoOp & \textbf{82.69} & 63.22 & 71.66 \\
    \rowcolor{tabhighlight}
    CoCoOp & 80.47 & 71.69 & \textbf{75.83} \\
    \bottomrule
    \end{tabular}
    \end{subtable}
    \vspace{1em}
    \begin{subtable}[t]{.3\textwidth}
    \centering
    \caption{ImageNet.}
    \begin{tabular}{l cc|c}
    \toprule
    & Base & New & H \\
    \midrule
    CLIP & 72.43 & 68.14 & 70.22 \\
    CoOp & \textbf{76.47} & 67.88 & 71.92\\
    \rowcolor{tabhighlight}
    CoCoOp & 75.98 & \textbf{70.43} & \textbf{73.10} \\
    \bottomrule
    \end{tabular}
    \end{subtable}
    ~
    \begin{subtable}[t]{.3\textwidth}
    \centering
    \caption{Caltech101.}
    \begin{tabular}{l cc|c}
    \toprule
    & Base & New & H \\
    \midrule
    CLIP & 96.84 & \textbf{94.00} & 95.40 \\
    CoOp & \textbf{98.00} & 89.81 & 93.73 \\
    \rowcolor{tabhighlight}
    CoCoOp & 97.96 & 93.81 & \textbf{95.84} \\
    \bottomrule
    \end{tabular}
    \end{subtable}
    ~
    \begin{subtable}[t]{.3\textwidth}
    \centering
    \caption{OxfordPets.}
    \begin{tabular}{l cc|c}
    \toprule
    & Base & New & H \\
    \midrule
    CLIP & 91.17 & 97.26 & 94.12 \\
    CoOp & 93.67 & 95.29 & 94.47 \\
    \rowcolor{tabhighlight}
    CoCoOp & \textbf{95.20} & \textbf{97.69} & \textbf{96.43} \\
    \bottomrule
    \end{tabular}
    \end{subtable}
    \vspace{1em}
    \begin{subtable}[t]{.3\textwidth}
    \centering
    \caption{StanfordCars.}
    \begin{tabular}{l cc|c}
    \toprule
    & Base & New & H \\
    \midrule
    CLIP & 63.37 & \textbf{74.89} & 68.65 \\
    CoOp & \textbf{78.12} & 60.40 & 68.13 \\
    \rowcolor{tabhighlight}
    CoCoOp & 70.49 & 73.59 & \textbf{72.01} \\
    \bottomrule
    \end{tabular}
    \end{subtable}
    ~
    \begin{subtable}[t]{.3\textwidth}
    \centering
    \caption{Flowers102.}
    \begin{tabular}{l cc|c}
    \toprule
    & Base & New & H \\
    \midrule
    CLIP & 72.08 & \textbf{77.80} & 74.83 \\
    CoOp & \textbf{97.60} & 59.67 & 74.06 \\
    \rowcolor{tabhighlight}
    CoCoOp & 94.87 & 71.75 & \textbf{81.71} \\
    \bottomrule
    \end{tabular}
    \end{subtable}
    ~
    \begin{subtable}[t]{.3\textwidth}
    \centering
    \caption{Food101.}
    \begin{tabular}{l cc|c}
    \toprule
    & Base & New & H \\
    \midrule
    CLIP & 90.10 & 91.22 & 90.66 \\
    CoOp & 88.33 & 82.26 & 85.19 \\
    \rowcolor{tabhighlight}
    CoCoOp & \textbf{90.70} & \textbf{91.29} & \textbf{90.99} \\
    \bottomrule
    \end{tabular}
    \end{subtable}
    \vspace{1em}
    \begin{subtable}[t]{.3\textwidth}
    \centering
    \caption{FGVCAircraft.}
    \begin{tabular}{l cc|c}
    \toprule
    & Base & New & H \\
    \midrule
    CLIP & 27.19 & \textbf{36.29} & \textbf{31.09} \\
    CoOp & \textbf{40.44} & 22.30 & 28.75 \\
    \rowcolor{tabhighlight}
    CoCoOp & 33.41 & 23.71 & 27.74 \\
    \bottomrule
    \end{tabular}
    \end{subtable}
    ~
    \begin{subtable}[t]{.3\textwidth}
    \centering
    \caption{SUN397.}
    \begin{tabular}{l cc|c}
    \toprule
    & Base & New & H \\
    \midrule
    CLIP & 69.36 & 75.35 & 72.23 \\
    CoOp & \textbf{80.60} & 65.89 & 72.51 \\
    \rowcolor{tabhighlight}
    CoCoOp & 79.74 & \textbf{76.86} & \textbf{78.27} \\
    \bottomrule
    \end{tabular}
    \end{subtable}
    ~
    \begin{subtable}[t]{.3\textwidth}
    \centering
    \caption{DTD.}
    \begin{tabular}{l cc|c}
    \toprule
    & Base & New & H \\
    \midrule
    CLIP & 53.24 & \textbf{59.90} & 56.37 \\
    CoOp & \textbf{79.44} & 41.18 & 54.24 \\
    \rowcolor{tabhighlight}
    CoCoOp & 77.01 & 56.00 & \textbf{64.85} \\
    \bottomrule
    \end{tabular}
    \end{subtable}
    ~
    \begin{subtable}[t]{.3\textwidth}
    \centering
    \caption{EuroSAT.}
    \begin{tabular}{l cc|c}
    \toprule
    & Base & New & H \\
    \midrule
    CLIP & 56.48 & \textbf{64.05} & 60.03 \\
    CoOp & \textbf{92.19} & 54.74 & 68.69 \\
    \rowcolor{tabhighlight}
    CoCoOp & 87.49 & 60.04 & \textbf{71.21} \\
    \bottomrule
    \end{tabular}
    \end{subtable}
    ~
    \begin{subtable}[t]{.3\textwidth}
    \centering
    \caption{UCF101.}
    \begin{tabular}{l cc|c}
    \toprule
    & Base & New & H \\
    \midrule
    CLIP & 70.53 & \textbf{77.50} & 73.85 \\
    CoOp & \textbf{84.69} & 56.05 & 67.46 \\
    \rowcolor{tabhighlight}
    CoCoOp & 82.33 & 73.45 & \textbf{77.64} \\
    \bottomrule
    \end{tabular}
    \end{subtable}
\end{table*}

Our approach is mainly evaluated in the following three problem settings: 1) generalization from base to new classes within a dataset (Section~\ref{sec:experiments;subsec:base2new}); 2) cross-dataset transfer (Section~\ref{sec:experiments;subsec:transfer_crossdatasets}); 3) domain generalization (Section~\ref{sec:experiments;subsec:dg}). All models used in our experiments are based on the open-source CLIP~\cite{radford2021learning}.\footnote{\url{https://github.com/openai/CLIP}.} Before discussing the results, we provide the details of the experimental setup below.

\paragraph{Datasets}
For the first two settings, i.e., base-to-new generalization and cross-dataset transfer, we use the 11 image recognition datasets as in Zhou et al.~\cite{zhou2021coop}, which cover a diverse set of recognition tasks. Specifically, the benchmark includes ImageNet~\cite{deng2009imagenet} and Caltech101~\cite{fei2004learning} for classification on generic objects; OxfordPets~\cite{parkhi2012cats}, StanfordCars~\cite{krause20133d}, Flowers102~\cite{nilsback2008automated}, Food101~\cite{bossard2014food} and FGVCAircraft~\cite{maji2013fine} for fine-grained classification; SUN397~\cite{xiao2010sun} for scene recognition; UCF101~\cite{soomro2012ucf101} for action recognition; DTD~\cite{cimpoi2014describing} for texture classification; and finally EuroSAT~\cite{helber2019eurosat} for satellite imagery recognition. For domain generalization experiments, we use ImageNet as the source dataset and four other variants of ImageNet that contain different types of domain shift as the target datasets, namely ImageNetV2~\cite{recht2019imagenet}, ImageNet-Sketch~\cite{wang2019learning}, ImageNet-A~\cite{hendrycks2021natural} and ImageNet-R~\cite{hendrycks2021many}.

Following Zhou et al.~\cite{zhou2021coop}, we randomly sample for each dataset a few-shot training set while using the original test set for testing. We only evaluate the highest shot number studied in Zhou et al.~\cite{zhou2021coop}, i.e., 16 shots, which is sufficient to justify our approach. For learning-based models, the results are averaged over three runs.

\paragraph{Baselines}
The direct rival to our approach is CoOp~\cite{zhou2021coop}, which essentially learns \emph{static} prompts (in comparison to our \emph{dynamic} prompts). The zero-shot method, i.e., CLIP~\cite{radford2021learning} is also compared, which is based on manual prompts. It is worth mentioning that the manual prompt for each dataset was intensively tuned using \emph{all classes in the test data}~\cite{radford2021learning}.

\paragraph{Training Details}
Our implementation is based on CoOp's code.\footnote{\url{https://github.com/KaiyangZhou/CoOp}.} Throughout the experiments, we use the best available vision backbone in CLIP, i.e., ViT-B/16. Zhou et al.~\cite{zhou2021coop} have suggested that a shorter context length and a good initialization can lead to better performance and stronger robustness to domain shift. Therefore, we fix the context length to 4 and initialize the context vectors using the pre-trained word embeddings of ``a photo of a'' for both CoOp and CoCoOp. Due to the instance-conditional design, our approach is slow to train and consumes much more GPU memory than CoOp. Therefore, to ensure the model can fit into a GPU and meanwhile reduce the training time, we train CoCoOp with batch size of 1 for 10 epochs. Such a limitation is discussed in more detail in Section~\ref{sec:limitations}.

\begin{figure*}[t]
    \centering
    \includegraphics[width=\textwidth]{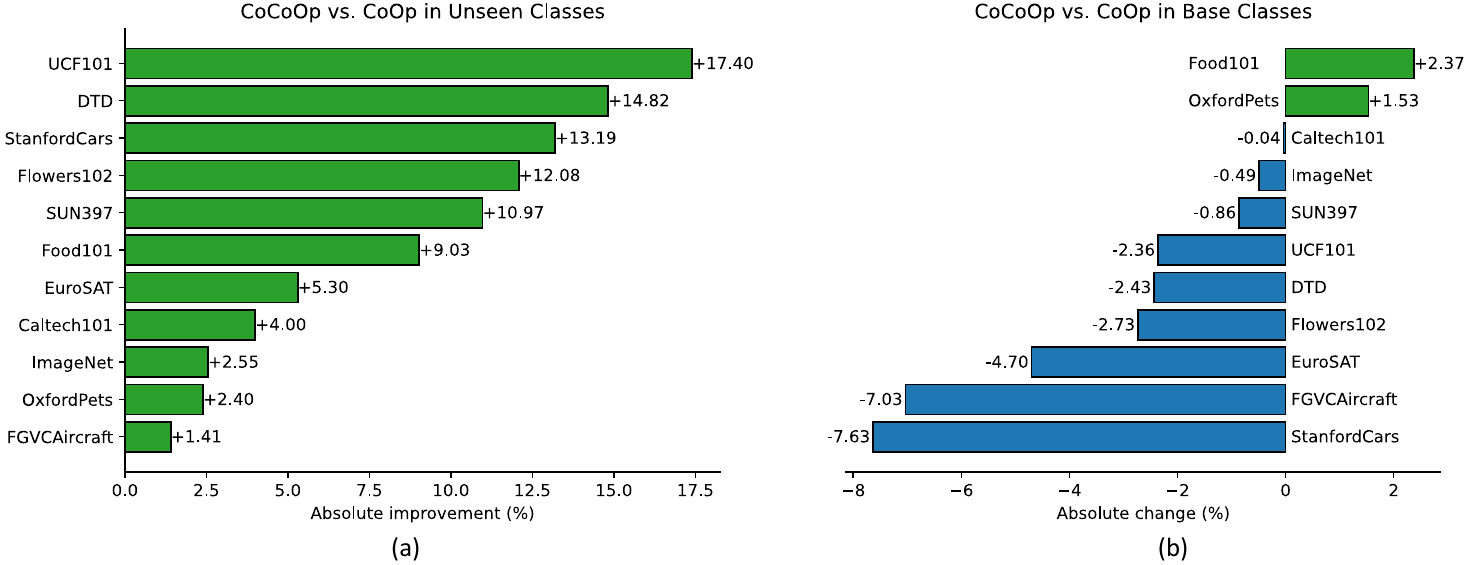}
    \caption{\textbf{Comprehensive comparisons of CoCoOp and CoOp in the base-to-new generalization setting}. (a) CoCoOp is able to gain consistent improvements over CoOp in unseen classes on all datasets. (b) CoCoOp's declines in base accuracy are mostly under 3\%, which are far outweighed by the gains in generalization.}
    \label{fig:coop_vs_ours_base2new}
\end{figure*}

\subsection{Generalization From Base to New Classes}
\label{sec:experiments;subsec:base2new}

Solving the weak generalizability problem of CoOp is the main focus in this research. On each of the 11 datasets, we split the classes equally into two groups, one as base classes and the other as new classes. Learning-based models, i.e., CoOp and CoCoOp, are trained using only the base classes while evaluation is conducted on the base and new classes \emph{separately} to test generalizability. The detailed results are shown in Table~\ref{tab:results_generalization}.

\paragraph{Failures of CoOp in Unseen Classes}
The split does not guarantee that the two class groups are equally difficult, as evidenced in CLIP's bumpy results: the base and new accuracy numbers are dramatically different.\footnote{For convenience, we refer to base accuracy as the performance in base classes; and similarly for new accuracy.} Nonetheless, CoOp's new accuracy is consistently much weaker than the base accuracy on nearly all datasets, leaving a huge gap of almost 20\% on average (82.69\% vs 63.22\%). Despite maintaining an advantage over CLIP in terms of average performance, CoOp's gains in the base classes are nearly zeroed out by the catastrophic failures in the new classes, highlighting the need to improve generalizability for learning-based prompts.

\paragraph{CoCoOp Significantly Narrows Generalization Gap}
As shown in Table~\ref{tab:results_generalization}(a), CoCoOp improves the accuracy in unseen classes from 63.22\% to 71.69\%, which largely reduces the gap with manual prompts. The results confirm that \emph{instance-conditional prompts are more generalizable}. A more detailed breakdown of per-dataset improvement is visualized in Figure~\ref{fig:coop_vs_ours_base2new}(a) where we observe more than 10\% increases in accuracy on 5 out of 11 datasets. Notably, on the challenging ImageNet dataset, CoCoOp's surge from 67.88\% to 70.43\% represents a non-trivial progress (the 70.43\% accuracy even surpasses CLIP's 68.14\%).

\begin{table*}[t]
    \tabstyle{5pt}
    \caption{\textbf{Comparison of prompt learning methods in the cross-dataset transfer setting}. Prompts applied to the 10 target datasets are learned from ImageNet (16 images per class). Clearly, CoCoOp demonstrates better transferability than CoOp. $\Delta$ denotes CoCoOp's gain over CoOp.
    }
    \label{tab:xd}
    \begin{tabular}{l c ccccccccccc}
    \toprule
    & Source & \multicolumn{11}{c}{Target} \\ \cmidrule(lr){2-2} \cmidrule(lr){3-13}
    & \rotbox{ImageNet} & \rotbox{Caltech101} & \rotbox{OxfordPets} & \rotbox{StanfordCars} & \rotbox{Flowers102} & \rotbox{Food101} & \rotbox{FGVCAircraft} & \rotbox{SUN397} & \rotbox{DTD} & \rotbox{EuroSAT} & \rotbox{UCF101} & \rotbox{\emph{Average}} \\
    \midrule
    CoOp~\cite{zhou2021coop} & \textbf{71.51} & 93.70 & 89.14 & 64.51 & 68.71 & 85.30 & 18.47 & 64.15 & 41.92 & \textbf{46.39} & 66.55 & 63.88 \\
    CoCoOp & 71.02 & \textbf{94.43} & \textbf{90.14} & \textbf{65.32} & \textbf{71.88} & \textbf{86.06} & \textbf{22.94} & \textbf{67.36} & \textbf{45.73} & {45.37} & \textbf{68.21} & \textbf{65.74} \\
    \midrule
    $\Delta$ & \hblue{-0.49} & \hgreen{+0.73} & \hgreen{+1.00} & \hgreen{+0.81} & \hgreen{+3.17} & \hgreen{+0.76} & \hgreen{+4.47} & \hgreen{+3.21} & \hgreen{+3.81} & \hblue{-1.02} & \hgreen{+1.66} & \hgreen{+1.86} \\
    \bottomrule
    \end{tabular}
\end{table*}

\begin{table*}[t]
    \tabstyle{7pt}
    \caption{\textbf{Comparison of manual and learning-based prompts in domain generalization}. CoOp and CoCoOp use as training data 16 images from each of the 1,000 classes on ImageNet. In general, CoCoOp is more domain-generalizable than CoOp.
    }
    \label{tab:dg}
    \begin{tabular}{l c ccccc}
    \toprule
    && Source & \multicolumn{4}{c}{Target} \\ \cmidrule(lr){3-3} \cmidrule(lr){4-7}
    & Learnable? & ImageNet & ImageNetV2 & ImageNet-Sketch & ImageNet-A & ImageNet-R \\
    \midrule
    CLIP~\cite{radford2021learning} & & 66.73 & 60.83 & 46.15 & 47.77 & 73.96 \\
    CoOp~\cite{zhou2021coop} & $\checkmark$ & \textbf{71.51} & \textbf{64.20} & 47.99 & 49.71 & 75.21 \\
    CoCoOp & $\checkmark$ & 71.02 & 64.07 & \textbf{48.75} & \textbf{50.63} & \textbf{76.18} \\
    \bottomrule
    \end{tabular}
\end{table*}

\paragraph{CoCoOp's Gains in Generalization Far Outweigh Losses in Base Accuracy}
In comparison to CoOp, performance drops in the base classes occur for CoCoOp on most datasets (see Figure~\ref{fig:coop_vs_ours_base2new}(b)). This is reasonable because CoOp optimizes specifically for base classes whereas \emph{CoCoOp optimizes for each instance in order to gain more generalization over an entire task}. But it is worth noting that on the 9 datasets where CoCoOp's base accuracy drops below CoOp's, most losses are under 3\% (precisely on 6 out of 9 datasets), which are far outweighed by the gains in unseen classes shown in Figure~\ref{fig:coop_vs_ours_base2new}(a); even for those where CoCoOp suffers the biggest losses, the boosts in generalization are mostly significant enough to turn the averages into positives, e.g., StanfordCars sees the worst base accuracy drop of -7.63\% but has the third-highest accuracy gain of +13.19\% in the new classes, which together bring a 5.56\% positive improvement for CoCoOp.

\paragraph{CoCoOp Is More Compelling Than CLIP}
When taking into account both the base and new classes, CoCoOp shows a gain of more than 4\% over CLIP (75.83\% vs 71.70), suggesting that \emph{instance-conditional prompts have a better potential in capturing more generalizable elements that are relevant for a recognition task}. Theoretically, learning-based prompts have a much higher risk of overfitting base classes than manual prompts. Therefore, CLIP is a strong competitor to beat in unseen classes. Different from CoOp, we obtain promising results for CoCoOp: the new accuracy is even better than CLIP's on 4 out of 11 datasets (i.e., ImageNet, OxfordPets, Food101 and SUN397) and not too far away from CLIP's on the rest except FGVCAircraft where the gap between manual and learning-based prompts is generally large. In the ablation study on context length, we find that FGVCAircraft benefits from longer context, which is aligned with the findings in Zhou et al.~\cite{zhou2021coop}. To close or even overturn the gaps between manual and learning-based prompts in unseen classes, more efforts are required and we hope the insights presented in this research can help the community tackle the generalizability issue in prompt learning.

\subsection{Cross-Dataset Transfer}
\label{sec:experiments;subsec:transfer_crossdatasets}

Having demonstrated CoCoOp's generalizability within a dataset, we further show that CoCoOp has the potential to transfer beyond a single dataset, which is a much more challenging problem because the fundamentals can be totally changed across different datasets (e.g., from object recognition to texture classification). We only consider prompt learning methods in this setting.

We compare CoCoOp with CoOp by transferring context learned from ImageNet, with all 1,000 classes used, to each of the other 10 datasets. The results are detailed in Table~\ref{tab:xd}. On the source dataset, the two models perform similarly. Whereas on the target datasets, CoCoOp mostly outperforms CoOp by a clear margin. Since the ImageNet classes mainly contain objects, as well as a fair amount of dog breeds, it is reasonable to see high accuracy for both models on the relevant target datasets including Caltech101 and OxfordPets. 

By comparison, the performance on other datasets with distant---and more fine-grained or specialized---categories is much lower, such as FGVCAircraft and DTD (containing various textures) where the accuracy numbers are well below 50\%. Nonetheless, CoCoOp exhibits much stronger transferability than CoOp on the two mentioned datasets as well as on most other fine-grained or specialized datasets.

\subsection{Domain Generalization}
\label{sec:experiments;subsec:dg}

Generalization to out-of-distribution data is a capability essential for machine learning models to succeed in practical applications~\cite{taori2020measuring,zhou2021domain}. Zhou et al.~\cite{zhou2021coop} have revealed that their learnable prompts are more robust than manual prompts to domain shift. We are also interested to know if instance-conditional prompts still maintain the advantages as in previous experiments.

Following Zhou et al.~\cite{zhou2021coop}, we evaluate CoCoOp's domain generalization performance by transferring the context learned from ImageNet to the four specially designed benchmarks. We also include the comparison with CLIP.
Table~\ref{tab:dg} shows the results. Both prompt learning methods clearly beat CLIP on all target datasets. Compared to CoOp, CoCoOp performs slightly worse on ImageNetV2 but better on the other three. The results confirm that \emph{instance-conditional prompts are more domain-generalizable}.

\subsection{Further Analysis}
\label{sec:experiments;subsec:further_analysis}

\begin{table}[t]
    \tabstyle{8pt}
    \caption{Recognition accuracy (average over 11 datasets) on a combination of base and new classes. The learnable models only have access to training data from base classes.
    }
    \label{tab:cls_incre_test}
    \begin{tabular}{l cc}
    \toprule
    & Learnable? & Accuracy \\
    \midrule
    CLIP~\cite{radford2021learning} & & 65.22 \\
    CoOp~\cite{zhou2021coop} & $\checkmark$ & 65.55 \\
    CoCoOp & $\checkmark$ & \textbf{69.13} \\
    \bottomrule
    \end{tabular}
\end{table}

\begin{figure}[t]
    \centering
    \includegraphics[width=\columnwidth]{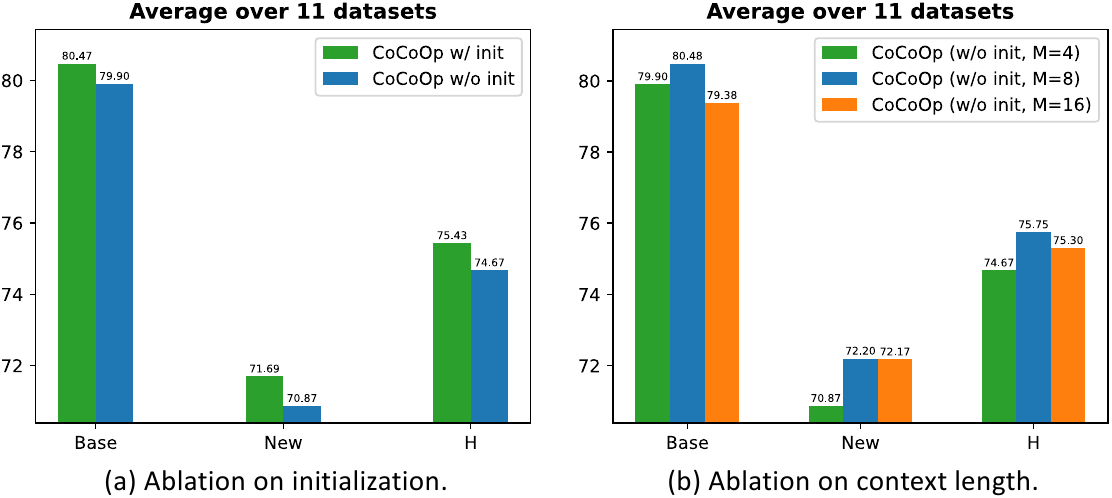}
    \caption{Ablation studies.}
    \label{fig:ablation_init_ctxlen}
\end{figure}

\paragraph{Class-Incremental Test}
We consider a practical problem scenario where the recognition targets originally composed of base classes are expanded to include completely new classes. This problem is relevant to the existing continual learning literature~\cite{parisi2019continual} but different in that \emph{the model here does not have access to any training data from new classes and needs to perform zero-shot recognition on them}. We compare CLIP, CoOp and CoCoOp using the 11 datasets. The average results are reported in Table~\ref{tab:cls_incre_test}. Clearly, CoOp loses competitiveness against CLIP as their performance is similar but the former needs training data. Again, CoCoOp beats the two competitors with a significant margin.


\paragraph{Initialization}
To understand the impact of initialization, we conduct an ablation study by comparing word embeddings-based initialization and random initialization while keeping all other parameters identical. For random initialization, we follow Zhou et al.~\cite{zhou2021coop} to sample from a zero-mean Gaussian distribution with 0.02 standard deviation. Figure~\ref{fig:ablation_init_ctxlen}(a) shows the base-to-new generalization results averaged over the 11 datasets, which suggest that a proper initialization is more beneficial to both the base and new classes. Note that the findings from Figure~\ref{fig:ablation_init_ctxlen} only represent the overall trend while each individual dataset might have a different result.


\paragraph{Context Length}
The ablation study on context length is also carried out in the base-to-new generalization setting. Following Zhou et al.~\cite{zhou2021coop}, we study 4, 8 and 16 context tokens. For fair comparison, we use random initialization for all context tokens. Figure~\ref{fig:ablation_init_ctxlen}(b) summarizes the results on the 11 datasets. The differences in the base classes are fairly small whereas in the new classes the models with a longer context length clearly perform better. From Figure~\ref{fig:ablation_init_ctxlen}(a) and (b) we observe that using 8 randomly initialized context tokens is marginally better than using 4 properly initialized context tokens, suggesting that a further boost might be possible if we initialize 8 context tokens with word embeddings.


\begin{table}[t]
    \tabstyle{3pt}
    \caption{CoCoOp (last row) vs a bigger CoOp on ImageNet.
    }
    \label{tab:vs_bigger_coop}
    \begin{tabular}{l c c c | c}
    \toprule
    Model & \# params & Base & New & H \\
    \midrule
    CoOp (ctx=4) & 2,048 & \textbf{76.47} & 67.88 & 71.92 \\
    CoOp (ctx=60) & 30,720 & 76.16 & 65.34 & 70.34 \\
    \rowcolor{tabhighlight}
    CoOp (ctx=4) + Meta-Net & 34,816 & 75.98 & \textbf{70.43} & \textbf{73.10} \\
    \bottomrule
    \end{tabular}
\end{table}

\paragraph{CoCoOp vs a Bigger CoOp}
Since CoCoOp introduces more parameters than CoOp, namely the Meta-Net, one might question if the improvements simply come from an increased learning capacity. To clear the doubt, we remove the Meta-Net part and increase the number of context tokens in CoOp to the maximum such that CoOp's and CoCoOp's sizes are similar. The results in Table~\ref{tab:vs_bigger_coop} show that increasing the parameter size is not the key.

\section{Limitations}
\label{sec:limitations}

The first limitation is about training efficiency: CoCoOp is slow to train and would consume a significant amount of GPU memory if the batch size is set larger than one. The reason is because CoCoOp is based on an instance-conditional design that requires for each image an independent forward pass of instance-specific prompts through the text encoder. This is much less efficient than CoOp that only needs a single forward pass of prompts through the text encoder for an entire mini-batch of any size.

The second limitation is that on 7 out of the 11 datasets (see Table~\ref{tab:results_generalization}), CoCoOp's performance in unseen classes still lags behind CLIP's, indicating that more efforts are needed from the community to fully close or overturn the gaps between manual and learning-based prompts.

\section{Discussion and Conclusion}
\label{sec:conclusion}

\begin{table*}[t]
    \tabstyle{9pt}
    \caption{Domain generalization results on DOSCO-2k, a recently proposed benchmark focusing on broader contextual domain shift. Among the three approaches, CoOp and its follow-up, CoCoOp, contain learnable components while CLIP here denotes the zero-shot model. Both CoOp and CoCoOp use four learnable context tokens initialized with the word embeddings of ``a photo of a''. Bold denotes the best performance on each dataset for a specific architecture.}
    \label{tab:dosco_2k}
    \begin{tabular}{l cccccccc}
    \toprule
    & P-Air & P-Cars & P-Ctech & P-Ins & P-Mam & P-Pets & P-UCF & \textit{Avg} \\
    \midrule
    \textbf{ResNet-50} & \\
    CLIP & 16.1 & 56.1 & 86.7 & 62.7 & 59.7 & 84.0 & 60.6 & 60.9 \\
    CoOp & \textbf{22.1} & \textbf{60.7} & 89.4 & 66.3 & 61.6 & 83.8 & \textbf{69.2} & 64.7 \\
    CoCoOp & 20.1 & 59.8 & \textbf{90.4} & \textbf{67.9} & \textbf{63.8} &  \textbf{87.6} & 69.1 & \textbf{65.5} \\
    \midrule
    \textbf{ResNet-101} & \\
    CLIP & 17.5 & 63.2 & 89.5 & 62.4 & 62.2 & 84.2 & 61.3 & 62.9 \\
    CoOp & \textbf{24.6} & \textbf{68.2} & 92.0 & 68.3 & 65.4 & 88.2 & \textbf{72.7} & \textbf{68.5} \\
    CoCoOp & 22.5 & 65.2 & \textbf{93.3} & \textbf{69.9} & \textbf{67.5} & \textbf{88.6} & 71.5 & 68.4 \\
    \midrule
    \textbf{ViT-B/32} & \\
    CLIP & 18.2 & 60.1 & 91.6 & 61.3 & 61.8 & 85.5 & 61.3 & 62.8 \\
    CoOp & \textbf{24.0} & \textbf{63.0} & 93.6 & 67.3 & 65.7 & \textbf{88.5} & \textbf{74.5} & \textbf{68.1} \\
    CoCoOp & 19.5 & 60.4 & \textbf{93.8} & \textbf{69.8} & \textbf{67.3} & \textbf{88.5} & 72.7 & 67.4 \\
    \midrule
    \textbf{ViT-B/16} & \\
    CLIP & 24.4 & 64.9 & 92.6 & 67.5 & 67.9 & 87.4 & 66.1 & 67.2 \\
    CoOp & \textbf{32.4} & \textbf{72.4} & 94.7 & 73.2 & 72.1 & 90.1 & \textbf{78.2} & \textbf{73.3} \\
    CoCoOp & 30.4 & 68.7 & \textbf{94.8} & \textbf{73.5} & \textbf{73.6} & \textbf{91.6} & 76.3 & 72.7 \\
    \bottomrule
    \end{tabular}
\end{table*}

Our research addresses an important issue that arises with the availability of large pre-trained AI models, i.e., how to adapt them to downstream applications. These models, also called foundation models~\cite{bommasani2021opportunities}, have received increasing attention from academia and industry in both the vision and NLP communities because they are so powerful in terms of their capabilities for diverse downstream tasks. However, foundation models are costly to pre-train in terms of data scale and compute resources; and typically contain an enormous number of parameters in order to develop sufficient capacity. For instance, the CLIP model~\cite{radford2021learning} based on ViT-B/16 used in our experiments has a whopping 150M parameter size. These factors together highlight the need for research of \emph{efficient adaptation methods for democratizing foundation models}.

Our studies, which follow the line of parameter-efficient prompt learning~\cite{zhou2021coop}, provide timely insights into the generalizability issue of static prompts, and more importantly, demonstrate that a simple design based on conditional prompt learning performs superbly in a variety of problem scenarios, including generalization from base to new classes, cross-dataset prompt transfer, and domain generalization.

In terms of future work, one direction is to further develop conditional prompt learning with potentially a more efficient implementation that can accelerate the training, as well as enhance generalizability. The cross-dataset transfer experiments indicate that instance-conditional prompts are more transferable---compared to static prompts---across tasks of varying natures. Therefore, it would be interesting to see if such an idea can scale to, e.g., bigger model size for the Meta-Net, larger-scale training images, and even heterogeneous training data mixed with different datasets.

\paragraph{Acknowledgements}
This work is supported by NTU NAP, and under the RIE2020 Industry Alignment Fund – Industry Collaboration Projects (IAF-ICP) Funding Initiative, as well as cash and in-kind contribution from the industry partner(s).

\appendix
\section*{Appendix}
\section{Results on DOSCO-2k}

\paragraph{DOSCO-2k}
The DOSCO (DOmain Shift in COntext) benchmark~\cite{zhou2022device} contains 7 image recognition datasets, which cover a wide range of classification problems, such as generic object recognition, fine-grained recognition on aircraft models, and action recognition. Unlike existing domain generalization datasets where the domain labels are manually defined and often limited to image style variations, DOSCO-2k focuses on broader contextual domain shift, which is automatically detected by a neural network pre-trained on the Places dataset~\cite{zhou2017places}. Following Zhou et al.~\cite{zhou2022device}, we use the 2k version where the training and validation splits in each dataset have 2,000 images in total (1,600 for training and 400 for validation).

\paragraph{Results}
We study three methods' domain generalization performance on DOSCO-2k: CLIP, CoOp and CoCoOp. All models are trained on the training set and the checkpoints with the best validation performance are used for final test in unseen domains. Table~\ref{tab:dosco_2k} shows the results of four different architectures. It is clear that the two learnable methods outperform the zero-shot method with a large margin, despite having only a small number of parameters to tune. CoCoOp beats CoOp on 4 out of 7 datasets but CoOp's average performance is higher. In summary, the results suggest that efficient adaptation methods like CoOp and CoCoOp have great potential in tackling transfer learning problems.

{\small
\bibliographystyle{ieee_fullname}
\bibliography{egbib}
}

\end{document}